%% file: 0-main.tex
\documentclass[letterpaper, 10 pt, conference]{ieeeconf}  

\IEEEoverridecommandlockouts                              

\usepackage{cite}
\usepackage{amsmath,amssymb,amsfonts}
\usepackage{algorithmic}
\usepackage{algorithm}
\usepackage{graphicx}
\usepackage{textcomp}
\usepackage{xcolor}
\usepackage[acronyms]{glossaries}
\usepackage{mathtools}
\usepackage[colorinlistoftodos,prependcaption]{todonotes}
\usepackage{regexpatch}
\usepackage{svg}
\usepackage{flushend} 
\usepackage{enumerate}
\makeatletter
\xpatchcmd{\@todo}{\setkeys{todonotes}{#1}}{\setkeys{todonotes}{inline,#1}}{}{}
\makeatother
\usepackage{tikz-network}
\usepackage{booktabs}
\usepackage{url}
\usepackage{caption}
\usepackage{subcaption}
\usepackage{comment}
\def\BibTeX{{\rm B\kern-.05em{\sc i\kern-.025em b}\kern-.08em
    T\kern-.1667em\lower.7ex\hbox{E}\kern-.125emX}}

\def\ie{\textit{i.e.},}

\def\sota{state-of-the-art}
\def\dataset{data set}
\def\groundtruth{ground truth}

\def\figvspace{\vspace{1em}}
\def\tabcapspace{\vspace{0em}}

\newcommand\numberthis{\addtocounter{equation}{1}\tag{\theequation}}

\newacronym{mcl}{MCL}{Monte-Carlo Localization}
\newacronym{ndt}{NDT}{Normal Distributions Transform}
\newacronym{cndt-om}{C-NDT-OM}{Clustered Normal Distributions Transform Occupancy Map}
\newacronym{ndt-mcl}{NDT-MCL}{Normal Distributions Transform Monte-Carlo Localization}
\newacronym[plural=NDT-OMs,firstplural=Normal Distributions Occupancy Maps (NDT-OM)]{ndt-om}{NDT-OM}{Normal Distributions Transform Occupancy Map}
\newacronym{ndt-se}{NDT-SE}{Semantically Enhanced Normal Distributions Transform}
\newacronym{d2d}{D2D}{Distribution-to-Distribution}
\newacronym{p2d}{P2D}{Point-to-Distribution}
\newacronym{ins}{INS}{Inertial Navigation System}
\newacronym{ned}{NED}{North-East-Down}
\newacronym{utm}{UTM}{Universal Transverse Mercator}
\newacronym{rmse}{RMSE}{Root Mean Squared Error}
\newacronym{ate}{ATE}{Absolute Trajectory Error}
\newacronym{rpe}{RPE}{Relative Pose Error}
\newacronym{hmm}{HMM}{Hidden Markov Model}
\newacronym{av}{AV}{autonomous vehicle}
\newacronym{ros}{ROS}{Robot Operating System}
\newacronym{gps}{GPS}{Global Positioning System}
\newacronym{gnss}{GNSS}{Global Navigation Satellite System}
\newacronym{sd-ndt-om}{SD-NDT-OM}{Semantic-Dynamic Normal Distributions Occupancy Map}
\newacronym{slam}{SLAM}{Simultaneous Localization and Mapping}
\newacronym{vslam}{vSLAM}{visual Simultaneous Localization and Mapping}
\newacronym{mle}{MLE}{maximum likelihood estimate}

\title{\LARGE \bf
Object-Oriented Grid Mapping in Dynamic Environments
}

\author{Matti Pekkanen, Francesco Verdoja, and Ville Kyrki 
\thanks{This work was supported by Business Finland, decision 9249/31/2021. We gratefully acknowledge the support of NVIDIA Corporation with the donation of the Titan Xp GPU used for this research.}
\thanks{M. Pekkanen, F. Verdoja and V. Kyrki are with School of Electrical Engineering,
        Aalto University, Espoo, Finland.
        {\tt\small \{firstname.lastname\}@aalto.fi}}%
}

\begin{document}
\bstctlcite{IEEEexample:BSTcontrol}
\maketitle
\thispagestyle{empty}
\pagestyle{empty}

\begin{abstract}
Grid maps, especially occupancy grid maps, are ubiquitous in many mobile robot applications. To simplify the process of learning the map, grid maps subdivide the world into a grid of cells whose occupancies are independently estimated using measurements in the perceptual field of the particular cell. However, the world consists of objects that span multiple cells, which means that measurements falling onto a cell provide evidence of the occupancy of other cells belonging to the same object. Current models do not capture this correlation and, therefore, do not use object-level information for estimating the state of the environment.
In this work, we present a way to generalize the update of grid maps, relaxing the assumption of independence. We propose modeling the relationship between the measurements and the occupancy of each cell as a set of latent variables and jointly estimate those variables and the posterior of the map.
We propose a method to estimate the latent variables by clustering based on semantic labels and an extension to the Normal Distributions Transform Occupancy Map (NDT-OM) to facilitate the proposed map update method. We perform comprehensive map creation and localization experiments with real-world \dataset{}s and show that the proposed method creates better maps in highly dynamic environments compared to \sota{} methods. Finally, we demonstrate the ability of the proposed method to remove occluded objects from the map in a lifelong map update scenario.
\end{abstract}

\input{1-intro}
\input{2-related-work}

\input{3-problem-statement}
\input{4-methods}
\input{5-1-experiments}
\input{5-2-results}

\input{6-conclusion}

\bibliographystyle{IEEEtran}
\bibliography{IEEEabrv, clean-full-abbr, ctrl}

\end{document}

%% file: 1-intro.tex
\section{Introduction}
\label{sec:intro}

Mapping is a central functionality in any mobile robot system, as accurate maps are required for foundational mobile robot capabilities such as localization and path planning. Grid maps, the most ubiquitous of which is the 2D occupancy grid map, discretize the environment into a grid of cells, which are assumed to be independent of each other such that only the measurements directly evaluating a cell as occupied or free are taken into account when estimating the occupancy of that cell. Cells are assumed to be independent to reduce the complexity of learning the map, and while it is known that this assumption is not entirely true, it is admissible when the environment is static. 

However, a single object often occupies more than one cell. This means that measurements from all cells occupied by the same object can be used to infer the occupancy of all object's cells. For example, if the object is partially occluded, the evidence from the visible cells should be used to infer the occupancy of the occluded cells. If the object has vacated and this is not done, the map will contain erroneously retained cells, which may cause problems in localization or path planning, as shown in Figure \ref{fig:header-img}.

\input{B-header-fig}

In this work, we formalize the theory of grid mapping with a relaxed cell-independence assumption and propose a way to use all measurements to update the cells in a grid map. This is achieved by jointly estimating the posterior occupancy of the map as well as a set of latent variables that describe a correspondence between the cells occupied by the same object, representing the object-oriented nature of the environment.

The benefit of this method is that more evidence is used for occupancy estimation by aggregating the evidence from the cells occupied by the same object, making the map adjust faster to the environment's dynamics. Additionally, we can infer the occupancy of occluded cells belonging to a partially occluded object by using the observations of the visible part of the object. This reduces the number of residual cells left by occlusions, improving the map quality.

In a series of experiments, we show the performance of the proposed method in a map creation task, where the number of erroneous residual cells is measured. Furthermore, we qualitatively demonstrate the performance in a lifelong map update scenario. Finally, we evaluate the quality of the produced maps in terms of localization accuracy.

The main contributions of this paper are:
\begin{enumerate}[i)]
    \item We formalize the theory for occupancy update on non-independent cells by jointly estimating the set of latent variables that describe the relationship between the measurements and the cells for general grid maps, yielding better quality maps in map creation and lifelong map update.
    \item We propose an extension to \gls{ndt-om} called \gls{cndt-om}, which uses the proposed method and estimates the latent variables by clustering the cells based on their semantic labels. We release the source code of the implementation.
\end{enumerate}

%% file: B-header-fig.tex
\begin{figure}[t]
    \centering
    \includegraphics[width=0.39\textwidth]{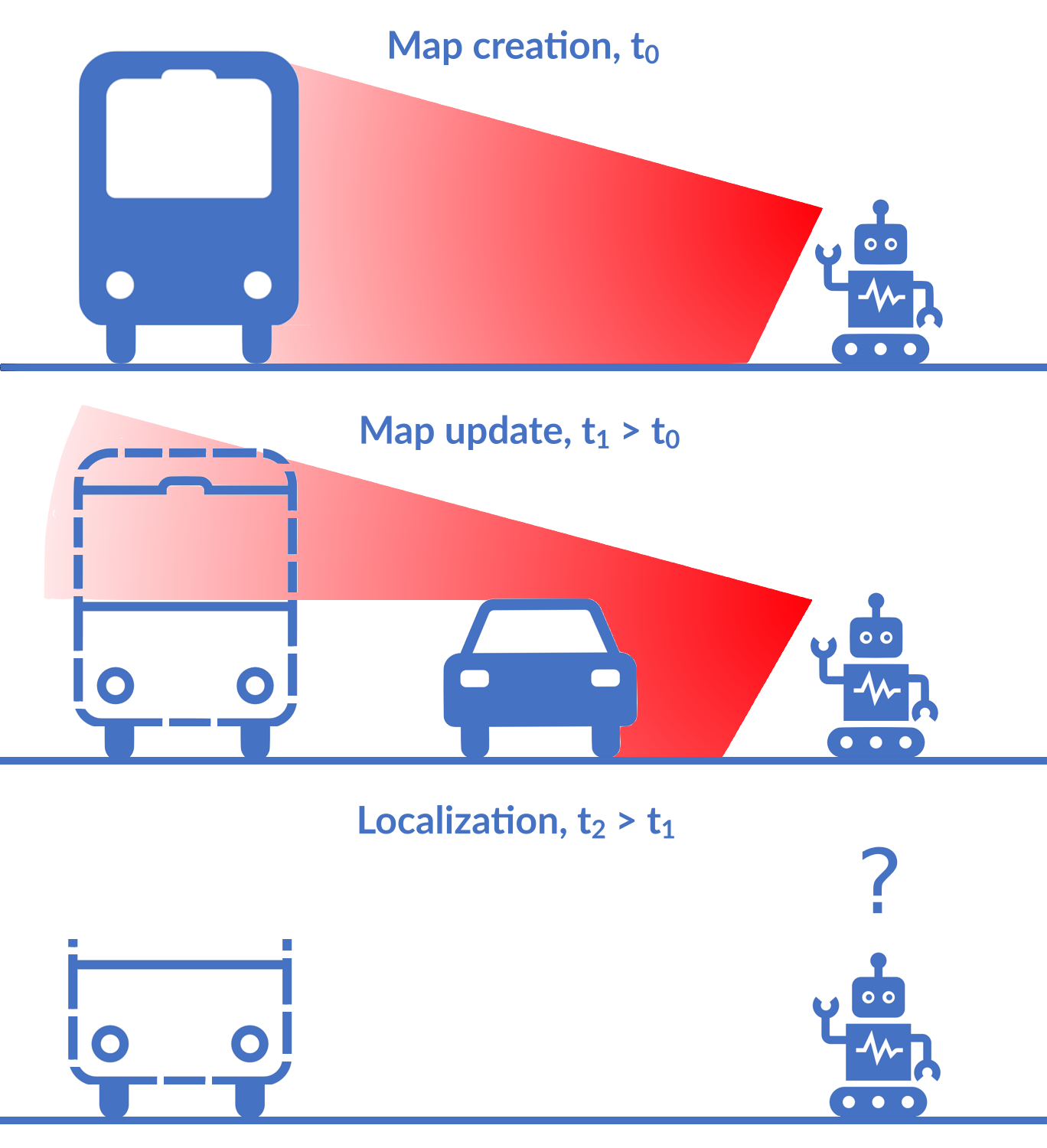}
    \caption{Occlusions and dynamic objects constitute a major source of mapping errors. When not treated properly, partial objects and other residual noise remain in the map.
    }
    \label{fig:header-img}
    \figvspace{}
\end{figure}

%% file: 2-related-work.tex
\section{Related work}
\label{sec:related}

\subsection{Mapping in dynamic environments}
\label{sec:dynamics}

Mapping in dynamic environments has been extensively studied. It was realized early on \cite{hahnel_map_2002} that dynamic objects violate the static world assumption that the maps incorporate. The most common solution is to detect and filter the dynamic measurements of laser scanners \cite{fan_dynamicfilter_2022, schmid_dynablox_2023}, recently, especially with neural networks \cite{mersch_receding_2022, zhou_motionbev_2023}. The use of semantic segmentation can aid the removal of dynamic objects, especially in \gls{vslam} \cite{ballester_dot_2021, singhFastSemanticAwareMotion2022}, but also increasingly with laser scanners as well \cite{chen_suma_2019, kim_rvmos_2022}. \glslocalunset{slam}

Other methods include continuous \gls{slam} instead of mapping, keeping temporary maps \cite{meyer-deliusTemporaryMapsRobust2010, morrisSimultaneousLocalizationPlanning2014} or maps of different timescales \cite{mitsouTemporalOccupancyGrid2007, valenciaLocalizationHighlyDynamic2014a}, modeling dynamics as \glspl{hmm} \cite{rappHiddenMarkovModelbased2016, meyer-deliusOccupancyGridModels2021a} or with frequency-based methods \cite{krajnikFreMEnFrequencyMap2017, krajnikWarpedHypertimeRepresentations2019a}. We point the interested reader towards \cite{sousaSystematicLiteratureReview2023a, beghdadi_comprehensive_2022} for more information.

While there are a plethora of ways to deal with dynamic objects before they are recorded in the map, we specifically want to extend the theory of grid maps, as they are the most widely used map representations in practical applications. Relaxing the independence assumption of the cells allows arbitrary formulation of the dependencies between the cells with the latent variables, which in turn allows a wider range of applications to extend the use of tried and tested grid maps in more challenging practical applications.

\subsection{Object-oriented maps}

The environment consists of background regions and discrete object instances. As the dynamic changes in maps typically occur on the object level \cite{fu_planesdf-based_2022} and are semantically consistent \cite{schmid_panoptic_2022}, it is natural to represent the map as a collection of objects. As in \cite{salas2013slampp}, maps that attempt to do that are commonly referred to in the literature as \emph{object-oriented maps}.

While multiple successful methods can track, add, and remove objects as a whole \cite{langer_robust_2020, grinvaldTSDFMultiObjectFormulation2021}, and represent their extent naturally, the probabilistic formulation of the occupancy of continuous space is still unsolved. On the other hand, grid maps can model the occupancy of space probabilistically. In this work, we propose a method to incorporate objects' extent into a grid map.

Graph sparsification methods used in graph \gls{slam} are similar to object-centric sub-mapping, except the sub-maps represent parts of the environment rather than object instances. However, the majority of these methods assume a static environment, and those that do not rely on replacing the measurements with newer ones \cite{walcott-bryant_dynamic_2012} or by naively combining the evidence \cite{einhorn_generic_2015}. In \cite{lazaro2018efficient}, a fusion method is presented, but it does not estimate the occupancy or probability of the update. Unlike the aforementioned methods, our work explicitly fuses the acquired evidence with the ability to remove vacated objects and add new objects.

\subsection{Object-oriented grid maps}

The two main methods of building object-oriented grid maps are to map the dynamics caused by objects or to combine object tracking and detection systems with mapping methods. An example of the former, and the closest method to ours, is \cite{jimenez_object-level_2023}, where the dynamic occupancy grid framework, proposed in \cite{nuss2018random}, was extended to include object clustering based on their dynamic properties to produce a 2.5D classified occupancy grid. However, unlike our work, the cells are assumed to be independent, and only measurements directly measuring the cell occupied or free are used for the cell occupancy estimation. In contrast, we use the whole set of measurements to estimate the cell occupancy.

The combination of object detection and tracking and \gls{slam} on occupancy grid was proposed in \cite{wang2007simultaneous} and further extended in \cite{gallagher_gatmo_2009} to track and update moving and movable objects. Unlike our work, the objects are stored in separate maps, and the cells are assumed to be independent within the object. In \cite{wang2020fusion}, the tracked dynamic objects are removed and added to the map after each step, removing the object trails. However, the cells are assumed to be independent, and the static and dynamic parts of the environment cannot overlap. Unlike that work, we can model the extent of the object and represent movable objects that can be updated in a lifelong scenario.

%% file: 3-problem-statement.tex
\section{Problem statement}
\label{sec:problem}

The occupancy mapping problem is defined as finding a point estimate for the posterior distribution $p(m_t | z_{0:t}, x_{0:t})$, where $m_t = \{ m^0_t, ..., m^n_t \}$ is the map at time $t$ and $m_t^i$ are the individual cells of the map, $x_{0:t} = \{x_0, ..., x_t\}$ is the sequence of the poses of the robot, and $z_{0:t} = \{z_0, ..., z_t\}$ the sequence of measurements.

Under the assumption that the cells are independent of each other, the map distribution can be solved as the product of its marginals. Additionally, it is generally assumed implicitly that only the measurements on whose perceptual field the cell is in---hereafter referred to as measurements \emph{falling onto} the cell---are used to update the cell. This yields the general formulation of the mapping problem as $\max_m p(m_t | z_{0:t}, x_{0:t}) = \max_m \prod_i p(m^i_t | z^i_{0:t}, x_{0:t})$, where $z^i_{0:t}$ is the set of measurements falling onto the cell $m^i$. This is depicted in Figure \ref{fig:standard-bayes}.

However, since the world consists of objects that span multiple cells, all measurements falling onto cells representing parts of the same object affect the occupancy of all cells occupied by the object. This means there exists a set of dependencies between the cells and measurements, and knowing these dependencies allows us to consider all measurements when estimating the occupancy of a single cell.

Let $d_t^{(i,j)}$ be a latent variable that models the effect on the cell $m^i$ of the measurements $z^j$, \ie{} those falling onto the cell $m^j$, at time $t$; also, let $\mathcal{D}_t$ be set of latent variables $d_t$ between all pairs of cells in the map. Using these definitions, we can improve our estimate of the occupancy of a cell $m^i$ by augmenting the measurement model by taking into account the whole set of measurements $p(m^i_t | \mathcal{D}_t, z_{0:t}, x_{0:t}) = p(m^i_t | z^i_{0:t}, d^{(i,i)}_{t}, ..., z^j_{0:t}, d^{(i,j)}_{t}, x_{0:t})$. Given the latent variables $\mathcal{D}_t$ are independent of each other, this can be solved as a product
\begin{align*}
    &p(m^i_t | \mathcal{D}_t, z_{0:t}, x_{0:t}) = \prod_j p(m^i_t | z^j_{0:t}, d^{(i,j)}_{t}, x_{0:t})\enspace,\numberthis
    \label{eq:measurement-model}
\end{align*}
which is depicted in Figure \ref{fig:proposed-bayes}.

To be able to use the augmented measurement model, we start by jointly estimating the occupancies and the latent variables
\begin{align*}
    &p(m_t, \mathcal{D}_t | z_{0:t}, x_{0:t}) = p(m_t | \mathcal{D}_t, z_{0:t}, x_{0:t}) \cdot p(\mathcal{D}_t | z_{0:t}, x_{0:t})\enspace, \numberthis
\end{align*}
and given that the latent variables $\mathcal{D}_t$ fully characterize the dependencies between each cell in the map, the cells of the map are conditionally independent of each other given $\mathcal{D}_t$; therefore, we can solve each cell independently
\begin{align*}
    &p(m_t, \mathcal{D}_t | z_{0:t}, x_{0:t}) =  \\
    &\prod_{i} p(m^i_t | \mathcal{D}_t, z_{0:t}, x_{0:t}) \cdot p(\mathcal{D}_t | z_{0:t}, x_{0:t})\enspace. \numberthis
\end{align*}

By marginalizing the latent variables, we acquire the posterior distribution of the map occupancy. The problem is to find a point estimate
\begin{align*}
    &\max_m \; p(m_t | z_{0:t}, x_{0:t}) = \\
    &\max_m \int p(m_t | \mathcal{D}_t, z_{0:t}, x_{0:t}) \cdot p(\mathcal{D}_t | z_{0:t}, x_{0:t}) \; d\mathcal{D}_t\enspace. \numberthis
    \label{eq:problem}
\end{align*}

\input{B-bayesian-network}

%% file: B-bayesian-network.tex
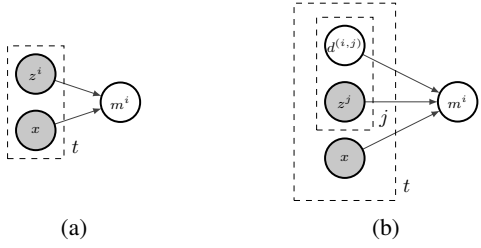
\begin{figure}[t]
    \centering
    \begin{subfigure}{0.2\textwidth}
        \centering
        \scalebox{0.75}{
            \begin{tikzpicture}
            \Vertex[x=0,y=0.5,Math,IdAsLabel,RGB,color={200,200,200},size=0.7]{x}
            \Vertex[x=0,y=1.5,Math,IdAsLabel,RGB,color={200,200,200},size=0.7]{z^i}
            \Vertex[x=1.5,y=1,Math,IdAsLabel,RGB,color={255,255,255},size=0.7]{m^i}
            \Edge[Direct,lw=0.5](x)(m^i)
            \Edge[Direct,lw=0.5](z^i)(m^i)
            \node[rectangle,draw,minimum width = 1cm, minimum height = 2cm,dashed] (r) at (0,1) {};
            \Text[x=0.7,y=0.2]{$t$}
             \node[rectangle,draw=none,minimum width = 1cm, minimum height = 2cm,dashed] (r) at (0,0.25) {};
            \end{tikzpicture}
        }
        \caption{}
        \label{fig:standard-bayes}
    \end{subfigure}
    \hfil
    \begin{subfigure}{0.2\textwidth}
        \centering
        \scalebox{0.75}{
            \begin{tikzpicture}
            \Vertex[x=0,y=0,Math,IdAsLabel,RGB,color={200,200,200},size=0.7]{x}
            \Vertex[x=0,y=1,Math,IdAsLabel,RGB,color={200,200,200},size=0.7]{z^j}
            \Vertex[x=0,y=2,Math,IdAsLabel,RGB,color={255,255,255},size=0.7]{d^{(i,j)}}
            \Vertex[x=2,y=1,Math,IdAsLabel,RGB,color={255,255,255},size=0.7]{m^i}
            \Edge[Direct,lw=0.5](x)(m^i)
            \Edge[Direct,lw=0.5](z^j)(m^i)
            \Edge[Direct,lw=0.5](d^{(i,j)})(m^i)
            \node[rectangle,draw,minimum width = 1cm, minimum height = 2cm,dashed] (r) at (0,1.5) {};
            \Text[x=0.7,y=0.7]{$j$}
            \node[rectangle,draw,minimum width = 1.8cm, minimum height = 3.5cm,dashed] (r) at (0,1) {};
            \Text[x=1.1,y=-0.5]{$t$}
            \end{tikzpicture}
        }
        \caption{}
        \label{fig:proposed-bayes}
    \end{subfigure}
    \caption{Inverse sensor model of the generic mapping method (a) and the proposed method (b). The vertices in gray are assumed to be known, while the vertices in white, \ie{} the map cells $m^i$ and the latent variables $d^{(i,j)}$, are unknown. We present the inverse sensor model instead of the usual likelihood model for clarity.}
    \figvspace{}
\end{figure}

%% file: 4-methods.tex
\section{Methods}
\label{sec:methods}

Marginalizing the latent variables $\mathcal{D}_t$ is intractable; therefore, we simplify the expression by assuming that the point estimate for the posterior is acquired by
\begin{align*}
    &\max_m \int p(m_t | \mathcal{D}_t, z_{0:t}, x_{0:t}) \cdot p(\mathcal{D}_t | z_{0:t}, x_{0:t}) \; d\mathcal{D}_t \simeq \\
    &\max_m  p(m_t | \mathcal{D}_t, z_{0:t}, x_{0:t}) \cdot \max_{\mathcal{D}} p(\mathcal{D}_t | z_{0:t}, x_{0:t})\enspace, \numberthis
    \label{eq:problem-solution}
\end{align*}

approximating the distribution of the $\mathcal{D}_t$ with its \gls{mle}, $\mathcal{D}_t \sim \delta(\mathcal{D}_t - \max_{\mathcal{D}} \mathcal{D}_t)$. This is sufficient in cases where $\mathcal{D}_t$ is distributed unimodally and can be estimated relatively well from the measurements, \ie{} there is no significant ambiguity in the estimate. Furthermore, as we are not trying to estimate the distribution of $m_t$, but rather its \gls{mle}, even if there would be multiple modes of $\mathcal{D}_t$, the less probable secondary modes would less likely affect the \gls{mle} of $m_t$. This approximation cannot be used to estimate the distribution $p(m_t | z_{0:t}, x_{0:t})$, as the approximation does not support multiple hypotheses of $\mathcal{D}_t$.

This splits the estimation problem into two parts. In Section \ref{sec:measurement-model} we propose a way to estimate $\max_m  p(m_t | \mathcal{D}_t, z_{0:t}, x_{0:t})$, and in Section \ref{sec:clustering} we propose a way to estimate $\max_{\mathcal{D}} p(\mathcal{D}_t | z_{0:t}, x_{0:t})$.

\subsection{Measurement model}
\label{sec:measurement-model}

The measurement model $p(m^i_t | \mathcal{D}_t, z_{0:t}, x_{0:t})$ is an extension to the standard inverse measurement model $p(m^i_t | z^i_{t}, x_{t})$ such that it takes into account the evidence gained from the other cells in the map, weighted by the latent variables $\mathcal{D}_t$, as shown in (\ref{eq:measurement-model}). The extended update term is implemented in log odds form as
\begin{align*}
    l(m^i_t) &=
    \sum_j \log \frac{p(m^i_t | z^j_{t}, x_{t})}{1 - p(m^i_t | z^j_{t}, x_{t})} \cdot d^{(i,j)}_{t} \\
    l(m^i_{0:t}) &= l(m^i_{0:t-1}) + l(m^i_t) \\
    \max_m p(m^i_t | \mathcal{D}_{t}, z_{0:t}, x_{0:t}) &= 1 - \frac{1}{1+\exp\{l(m^i_{0:t})\}}\enspace, \numberthis
\end{align*}
which implements a binary Bayes filter yielding the maximum a posteriori estimate of the cell occupancy probability. It is noteworthy that when $\mathcal{D}_t = \{ d^{(i,j)}_t = 0 : m^i_t m^j_t \in m_t, j \neq i \} \cup \{ d^{(i,i)}_t = 1 : m^i_t \in m_t \}$, the problem resolves back to the original mapping problem with independent cells.

\subsection{Estimation of the latent variables}
\label{sec:clustering}

The intuition behind the latent variables $\mathcal{D}_t$ is that the environment consists of objects that span multiple cells, and a positive latent variable between two cells reflects that the same object is occupying both cells. This makes the relationship symmetric, and therefore we model the environment as \emph{clusters} $c \subseteq m$ representing the objects in the environment. All the elements in a cluster are cells that are dependent on each other, \ie{} $\forall m^i m^j \in c, d^{(i,j)} > 0$.

A map with $n$ cells has $n^2$ latent variables in $\mathcal{D}_t$. Therefore, instead of estimating the pairwise latent variables between each two cells in the map, we estimate the \emph{membership} $\delta^i$ of the cell $m^i$ to a cluster $c$. We assume that clusters form a partition of the cells of the map, therefore the number of clusters $n_c \leq n$. From this, the entire set of latent variables $\mathcal{D}_t$ is recoverable by choosing $\forall m^j \in c, d^{(j,i)} = \delta^i$  for all clusters.

While any clustering approach can be employed to estimate $\mathcal{D}_t$, we implement it as a region growing algorithm \cite{adams1994seeded} based on the cell semantics. When the map is initialized after the first measurement---a laser point cloud---each voxel in the map is initialized as a seed cluster. Then, after $N$ scans are added to the map, region growing steps are performed. We used $N=1$, as the clustering does not introduce significant computational overhead. In the growing step,  the labels of all neighboring clusters in the map are compared: if the labels are the same, the clusters are joined; otherwise, they are not. This step is iterated until no more clusters can be joined or a maximum iteration limit is reached. A similarity metric based on the Kullback-Leibler divergence between the label distributions was tested, but the simpler metric proved to be sufficiently accurate while computationally lighter.

The membership, $\delta^i$, is estimated using Pearson's $\chi^2$ test, determining whether two categorical distributions are independent.

In an occupancy grid map, each measurement falling onto a cell provides evidence of the cell being either occupied or empty. Therefore, the measurements form a categorical distribution with two categories \emph{occupied} and \emph{empty}, and the number of measurements of each category for a cell $m^i$ are denoted as $o^i$ and $e^i$, respectively. The same quantities over the whole cluster $c$ can also be calculated, but the evidence is weighed according to the membership of that cell to the cluster. Therefore the evidence of \emph{occupied} over the cluster $c$ is
\begin{align}
    o^c &= \sum_{m^j \in c} \delta^j \cdot o^j\enspace,
\end{align}
and the evidence of \emph{empty} $e^c$ is computed similarly. This allows us to estimate the membership $\delta^i$ of the cell $m^i$ in the cluster $c$
\begin{align}
    \delta^i &= \chi^2 \bigg(\frac{(o^i - o^c)^2}{o^c} + \frac{(e^i - e^c)^2}{e^c} \bigg)\enspace.
\end{align}

If the evidence of the cell and the cluster align very well, the cell is likely a member of the cluster. However, if the evidence differs significantly, it is likely that the cell is no longer a member of the cluster. This may happen for various reasons, such as the original cluster moving away and a new object occupying the cell.

The \gls{ndt-om} measurement model $p(m^i_t | z_{t}, x_{t})$ is used in this work, which is described in detail in \cite{saarinen_ndtom_2013}. Instead of directly counting the rays that pass through the voxel as empty observations, in \gls{ndt-om}, the probability $p(x_{ML} | z_i)$ is estimated whether the ray passed through the distribution or not. A threshold $p_{th} = 0.5$ is set for the probability whether the measurement is counted as \emph{occupied} or \emph{empty}. Additionally, \gls{ndt-om} contains a scaling factor $\eta \in [0, 0.5]$ that controls the adaptation rate of the map. Following common practice, $\eta = 0.2$ is used unless otherwise stated.

\subsection{C-NDT-OM}

In this work, we propose an extension to \gls{ndt-om} \cite{saarinen_ndtom_2013}, the \gls{cndt-om}. \gls{ndt-om} divides the environment into a regular 3D grid of cells, with each cell $m^i$ represented by a 3D normal distribution $\mathcal{N}(\mu^i, \Sigma^i)$, number of measurements $N^i$ and the probability of the cell being occupied $p(m^i|z_{0:t})$.

In \gls{cndt-om}, the cell definition is extended by storing the cluster index $c$, cluster membership $\delta^i$, and the categorical distribution of semantic labels $p(l|\vec{p})$, yielding the augmented cell definition of $m^i = \{ \mu^i, \Sigma^i, N^i, p(m^i|z_{0:t}), p(l|\vec{p}), c, \delta^i\}$. This gives us the definition of the \gls{cndt-om} as a set of cells $m_{cndt} = \{ m^i, ..., m^n \}$.

%% file: 5-1-experiments.tex
\section{Experiments}
\label{sec:experiments}

The two main questions we want to answer with the experiments are:
\begin{enumerate}
    \item Does the number of residual cells caused by dynamic objects reduce using the proposed object-oriented map update method?
    \item Does the number of residual cells caused by movable objects reduce in a lifelong localization scenario using the proposed object-oriented map update method?
\end{enumerate}

To answer these questions, we compared the proposed mapping method to the \sota{} baseline method, \gls{ndt-om}, in a series of experiments\footnote{\url{https://github.com/aalto-intelligent-robotics/lamide}}.

In the first experiment (Section \ref{sec:mapping-exp}), we created 11 maps with each method and counted all the dynamic cells in the map as a metric of how well the methods can cope with dynamic objects. In the second experiment, we demonstrated the capabilities of the proposed mapping method on a lifelong update scenario (Section \ref{sec:lifelong-exp}).

Additionally, (Section \ref{sec:loc-exp}), we evaluated the impact of the map quality on the localization accuracy by creating a map with each method, performing seven localization experiments on each map, and evaluating the localization accuracy.

\subsection{Mapping experiment}
\label{sec:mapping-exp}

The Semantic Kitti \dataset{} \cite{behley2019iccv, Geiger2013IJRR} was used in the mapping experiment as the \dataset{} provides \groundtruth{} semantic labels. The \dataset{} contains measurements from a Velodyne HDL-64E laser and given \groundtruth{} poses used for map building.

For both the baseline \gls{ndt-om} and the proposed mapping method, we employed the NDT Fusion method \cite{stoyanov_normal_2013} using \groundtruth{} poses provided by the \dataset{}. Voxel size of $0.6$~m and submaps with $(x,y,z)$-dimensions of $(200, 200, 20)$m were used. Maps were created with each method from all of the sequences of the \dataset{} where \groundtruth{} poses were provided, \ie{} sequences 00--10. Thus, in total, 11 maps were created per method.

The number of dynamic objects present in the scenes was measured, and the sequences were split into two groups: sequences with high dynamics and sequences with low dynamics.

After the maps were created, the number of dynamic cells, \ie{} cells with a dynamic majority label, \ie{} labels 252--259, were counted. This measures how well the map update can remove dynamic objects from the maps. Ideally, there would be a small number of dynamic cells, as the dynamic objects constitute a very small portion of the map. However, as the dynamic objects move, the cells previously occupied by the object remain occupied and must be emptied. Therefore, fewer dynamic cells indicate the increased ability to remove the trails left by the dynamic objects in the map. As the sequences are of different lengths, the number of dynamic cells is divided by the length of the sequence, providing the final measure: the density of dynamic cells.

\input{B-kitti-table}

\subsection{Lifelong map update experiment}
\label{sec:lifelong-exp}

The Oxford Radar RobotCar \dataset{} \cite{RobotCarDatasetIJRR, RadarRobotCarDatasetICRA2020} was used in the lifelong update experiment. This \dataset{} was selected as it consists of multiple traversals along the same route, allowing objects to move between sequences. It requires a lifelong map update, where the scans from later sequences are used to update the original map.

The \dataset{} contains data from two Velodyne 32E lasers, and only the measurements from the left laser were used, as they were sufficient for the experiment. The semantic segmentation of the laser point clouds was obtained using RandLA-net \cite{hu2019randla}, with a pre-trained model provided by the authors. The map creation method is the same as presented in Section \ref{sec:mapping-exp}.

Sequence 1 was chosen for map building, and sequences 2 and 6 were selected for the map update. This allows us to evaluate the quality of the methods in a lifelong map update. We compared the methods with the nominal value of $\eta = 0.2$ and the maximum adaptation rate $\eta = 0.5$.

As there are no readily available metrics to evaluate the quality of the map, we present a qualitative demonstration of the capabilities of our method compared to the baseline.

\subsection{Localization experiment}
\label{sec:loc-exp}

The Oxford Radar RobotCar \dataset{} was used in the localization experiment. Eight sequences were chosen from the \dataset{} containing over 265k laser scans. Sequence 1 was selected for map building, and sequences 2--8 for localization. The first sequence from each day of the \dataset{} was selected, except for sequences 5 and 6, where the first sequences suffered from prolonged losses of \gls{gnss} fix, making the \groundtruth{} unreliable.

Maps were created as presented in Section \ref{sec:mapping-exp}, and the sensor setup, semantic segmentation, map creation, and localization methods are the same as presented in Section \ref{sec:lifelong-exp}.

\gls{ndt-mcl} \cite{saarinen_mcl_2013} was used as a localization method in the experiment using the same motion model as in the original paper. The \gls{ins} was used as odometry. Localization was initialized around the known initial pose $x_0$ with uniform distribution with dimensions $[-20, 20]$m on $xy$-axes and $[0, 2\pi]$rad in $\psi$. All of the experiments were run at $0.2$ of real-time. The estimated pose was stored at each time step, as well as the \groundtruth{}, which were compared in terms of \gls{ate} \cite{sturm_benchmark_2012}.

The implementations of \gls{ndt-om} fusion and the \gls{ndt-mcl} were based on \cite{graph_map, velo_cloud, ndt_core, ndt_tools}.

\input{B-kitti-table2}

%% file: B-kitti-table.tex
\begin{table}[t]
    \caption{The results of the mapping experiment over the Kitti sequences with high dynamics in terms of the density of dynamic voxels in each map.}
    \tabcapspace{}
    \begin{center}
        \setlength\tabcolsep{6pt}
        \scalebox{1.0}{
                \begin{tabular}{rrr}
                    \toprule
                    \textbf{Sequence} & \textbf{ \gls{ndt-om} } & \textbf{ Our method }  \\
                    \midrule
                    04 & 4.582 & \textbf{2.893}\\
                    01 & 3.426 & \textbf{2.298}\\
                    \midrule
                    mean & 4.004 & \textbf{2.595}\\
                    \bottomrule
                \end{tabular}
        }
        \setlength\tabcolsep{6pt}
        \label{tab:kitti-results-dynamic}
    \end{center}
    \figvspace{}
\end{table}

%% file: B-kitti-table2.tex
\begin{table}[t]
    \caption{The results of the mapping experiment over the Kitti sequences with low dynamics in terms of the density of dynamic voxels in each map.}
    \tabcapspace{}
    \begin{center}
        \setlength\tabcolsep{6pt}
        \scalebox{1.0}{
                \begin{tabular}{rrr}
                    \toprule
                    \textbf{Sequence} & \textbf{ \gls{ndt-om} } & \textbf{ Our method }\\
                    \midrule
                    07 & \textbf{0.508} & 0.831  \\
                    10 & \textbf{0.308} & 0.539  \\
                    09 & 0.478 & \textbf{0.325}  \\
                    08 & \textbf{0.301} & 0.344  \\
                    05 & \textbf{0.182} & 0.185  \\
                    03 & 0.176 & \textbf{0.138}  \\
                    02 & \textbf{0.078} & 0.103  \\
                    00 & \textbf{0.073} & 0.094  \\
                    06 & \textbf{0.037} & 0.073  \\
                    \midrule
                    mean & \textbf{0.238} & 0.293 \\
                    \bottomrule
                \end{tabular}
        }
        \setlength\tabcolsep{6pt}
        \label{tab:kitti-results-static}
    \end{center}
    \figvspace{}
\end{table}

%% file: 5-2-results.tex
\subsection{Results}
\label{sec:results}

\subsubsection{Mapping experiment}
\label{sec:mapping-results}

The results of the mapping experiment in high dynamic scenarios are presented in Table \ref{tab:kitti-results-dynamic}. Sequences 01 and 04 of the \dataset{} contain considerably more dynamic objects than the other sequences, which can be seen from the dynamic densities. Our method clearly outperforms \gls{ndt-om} in highly dynamic environments, resulting in fewer residual cells from dynamic objects. The results support the proposition that our method yields benefits in highly dynamic environments.

In cases with few dynamic objects, presented in Table \ref{tab:kitti-results-static}, \gls{ndt-om} performs slightly better on average. However, the differences are minor and inconsistent over all sequences. This is likely due to the conservative nature of the membership estimation. The membership of the cells that were occupied by a moving object takes some empty measurements to clear. Especially in cases where the submap changes, some of the voxels might remain incorrectly part of a dynamic object. A weighting parameter for the membership estimation could be introduced to remedy this problem.

\input{B-qualitative-speed}
\input{B-qualitative-occlusion}
\input{B-qualitative-fail}

\subsubsection{Lifelong map update experiment}

From the qualitative results of the lifelong experiment presented in Figures \ref{fig:qualitative-maps-speed} and \ref{fig:qualitative-maps-occlusion}, we can see that our method is capable of delivering the intended benefit: when parts of the object are occluded, the cluster-based update can correctly remove the occluded parts of the object along with the observed removal of the cells (Figure \ref{fig:occlusion-cl}). Another key difference is speed (Figure \ref{fig:qualitative-maps-speed}): our method can remove the vehicle after only four scans, whereas even after all scans measuring that vehicle before moving away  (\ie{} 150 updates), with \gls{ndt-om}, parts of the vehicle are still in the map.

Moreover, the effect of adaptation parameter $\eta$ is demonstrated in Figure \ref{fig:qualitative-maps-speed}. When using the maximum adaptation rate of $\eta = 0.5$, the \gls{ndt-om} eventually removes the vehicle after 120 scans; our method outperforms \gls{ndt-om} after only two scans. Our method can consistently remove the vehicle with fewer scans and lower values of $\eta$. This demonstrates that our method improves upon the capabilities of \gls{ndt-om}.

Our method's most common failure mode is the erroneous joining of disjoint clusters of the same label due to their proximity. However, in environments with sufficient static structure, the false removal of semi-static structures is considered less problematic than the false matching of the objects. This phenomenon can be seen in a tightly occupied parking lot in Figure \ref{fig:qualitative-maps-failure}, where all cars are clustered together and removed. A more advanced clustering method would limit this behavior.

\subsubsection{Localization experiment}
\label{sec:localization-results}

The results of the localization experiment, presented in Figure \ref{fig:localization-results}, show that our method has a lower overall sum, median, and mean \gls{ate} over all of the sequences. While the differences are negligible, they indicate that our method does not reduce the localization accuracy but could address some of the failure modes of \gls{ndt-om} in dynamic environments. Overall, the traversed paths are long and do not provide sufficiently difficult environments where our method could provide meaningful improvements over \gls{ndt-om}.

In environments where \gls{ndt-om} performs well (\ie{} 4, 6, 8), our method does not yield benefits but instead introduces errors due most likely to the noisy semantic labels. We expect that in the presence of higher-quality labels, the performance of our method will increase.

\input{B-localization-results}

In the data set, there are several difficult areas where both of the methods struggle. These localization errors raise the ATE to higher than expected from the method. The ATE for both methods could likely be reduced by parameter tuning. Therefore, it is likely that the low localization accuracy masks some of the benefits of our method. However, for the comparison of the methods, the relative change in ATE is more important than the absolute values.

%% file: B-qualitative-speed.tex
\newcommand\imageScale{0.08}
\newcommand\divOne{0.162}

\begin{figure*}[ht]
    \begin{subfigure}[t]{\divOne\textwidth}
        \centering
        \scalebox{\imageScale}{\includegraphics{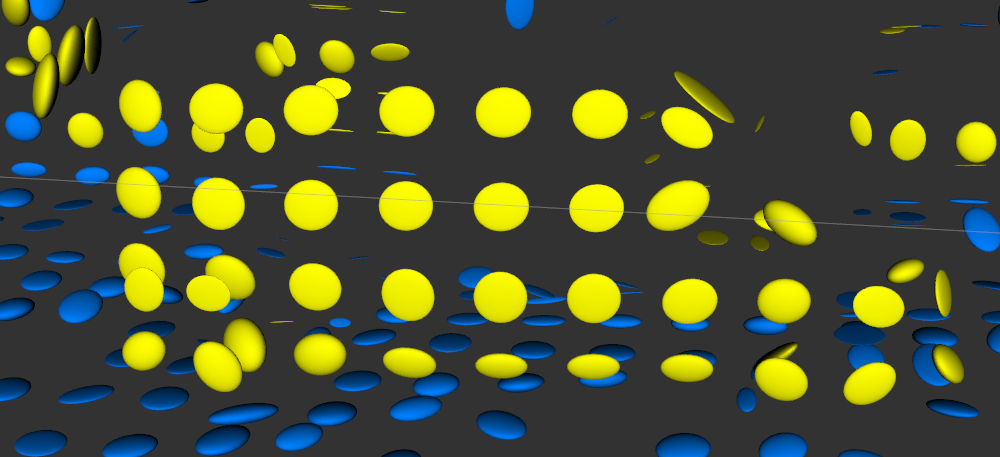}}
        \caption{Before the update}
        \label{fig:gt-car1}
    \end{subfigure}
    \begin{subfigure}[t]{\divOne\textwidth}
        \centering
        \scalebox{\imageScale}{\includegraphics{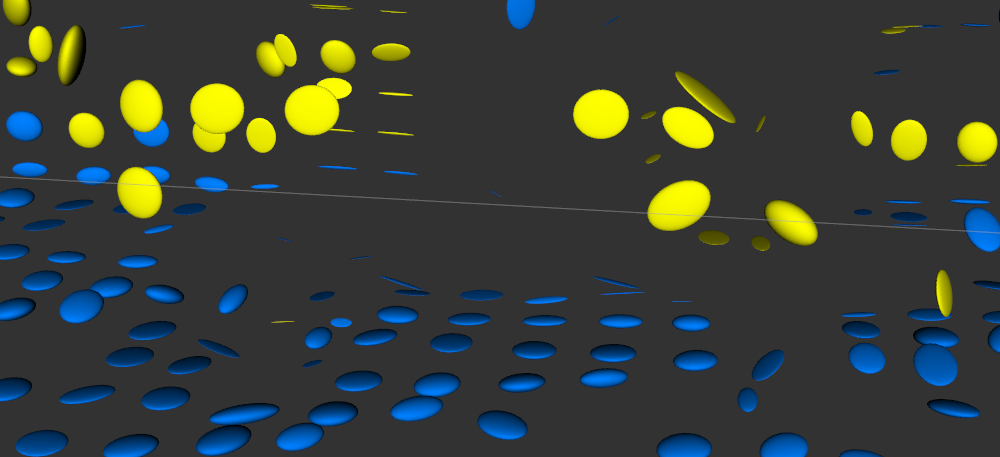}}
        \caption{NDT-OM,\\ $\eta = 0.2$, 150 scans}
        \label{fig:om-02-n-car1}
    \end{subfigure}
    \begin{subfigure}[t]{\divOne\textwidth}
        \centering
        \scalebox{\imageScale}{\includegraphics{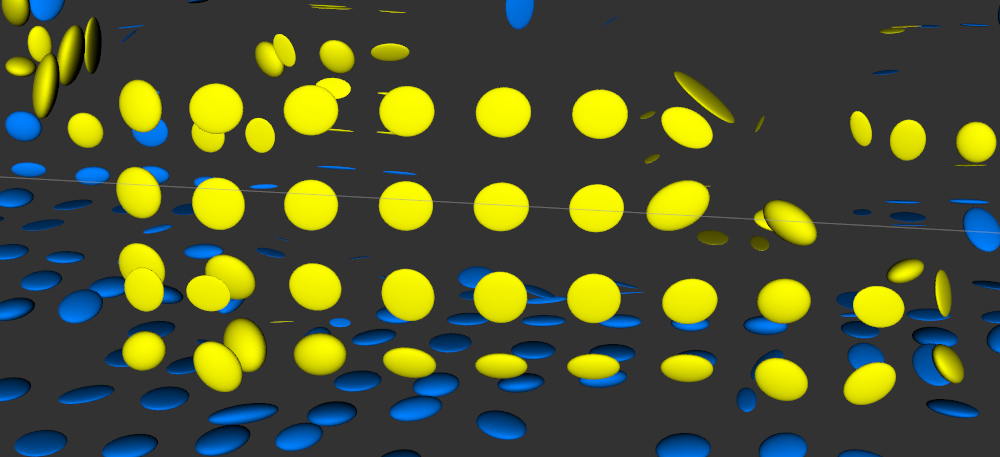}}
        \caption{NDT-OM,\\ $\eta = 0.5$, 2 scans}
        \label{fig:om-05-2-car1}
    \end{subfigure}
    \begin{subfigure}[t]{\divOne\textwidth}
        \centering
        \scalebox{\imageScale}{\includegraphics{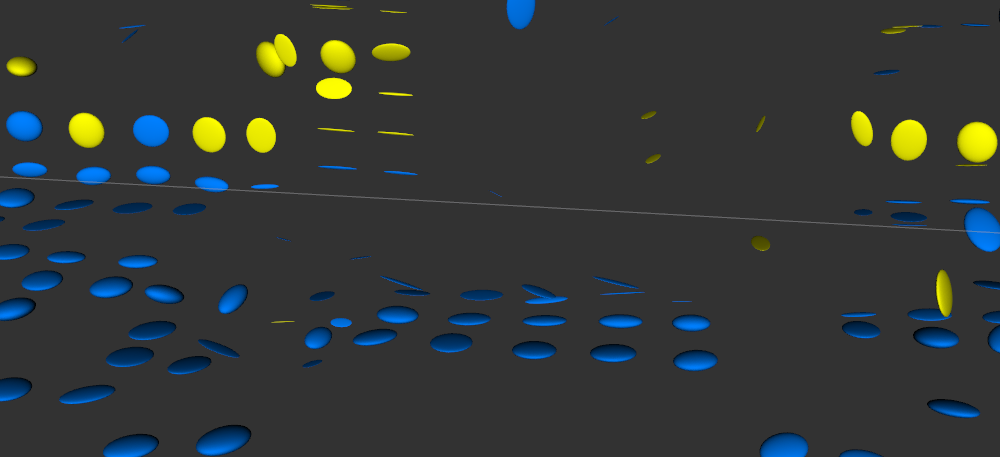}}
        \caption{NDT-OM,\\ $\eta = 0.5$, 120 scans}
        \label{fig:om-05-n-car1}
    \end{subfigure}
    \begin{subfigure}[t]{\divOne\textwidth}
        \centering
        \scalebox{\imageScale}{\includegraphics{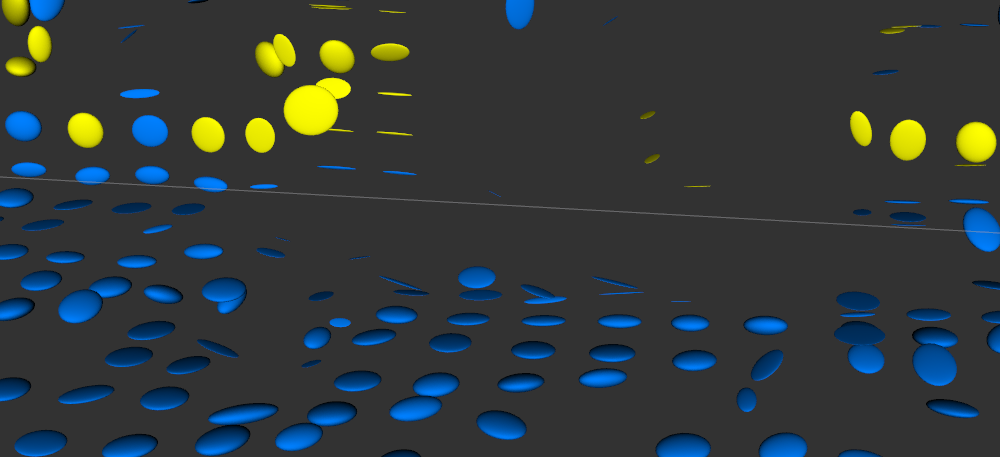}}
        \caption{Our method,\\ $\eta = 0.2$, 4 scans}
        \label{fig:cl-02-4-car1}
    \end{subfigure}
    \begin{subfigure}[t]{\divOne\textwidth}
        \centering
        \scalebox{\imageScale}{\includegraphics{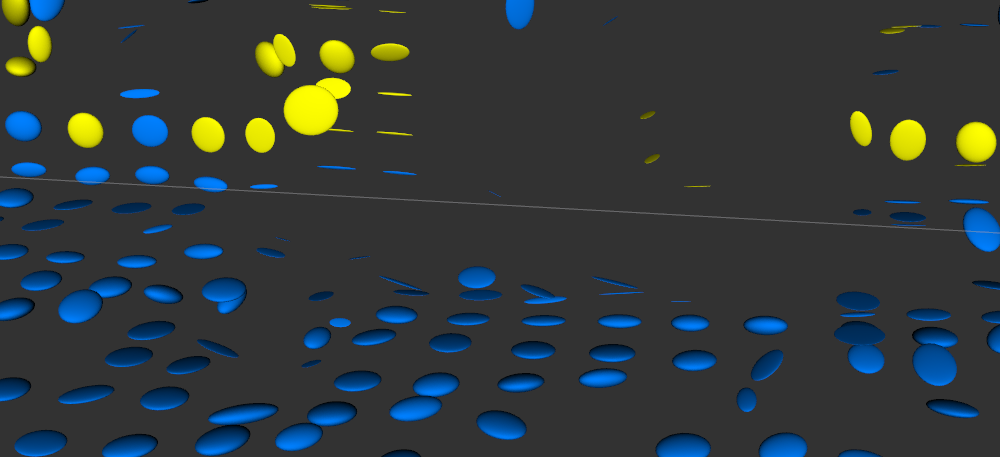}}
        \caption{Our method,\\ $\eta = 0.5$, 2 scans}
        \label{fig:cl-05-2-car1}
    \end{subfigure}
    \caption{The initial map (a) is updated in the lifelong scenario with \gls{ndt-om} (b) - (d), and our method (e) and (f). Our method outperforms \gls{ndt-om} in update speed. }
    \label{fig:qualitative-maps-speed}
    \figvspace{}
\end{figure*}

%% file: B-qualitative-occlusion.tex
\newcommand\imageScaleThree{0.08}
\newcommand\divThree{0.155}

\begin{figure}[ht]
    \begin{subfigure}[t]{\divThree\textwidth}
        \centering
        \scalebox{\imageScaleThree}{\includegraphics{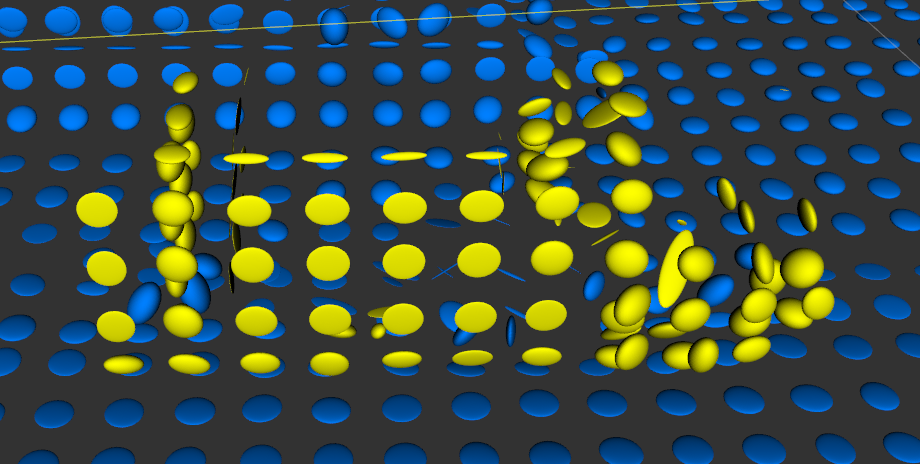}}
        \caption{Before the update}
        \label{fig:occlusion-gt}
    \end{subfigure}
    \begin{subfigure}[t]{\divThree\textwidth}
        \centering
        \scalebox{\imageScaleThree}{\includegraphics{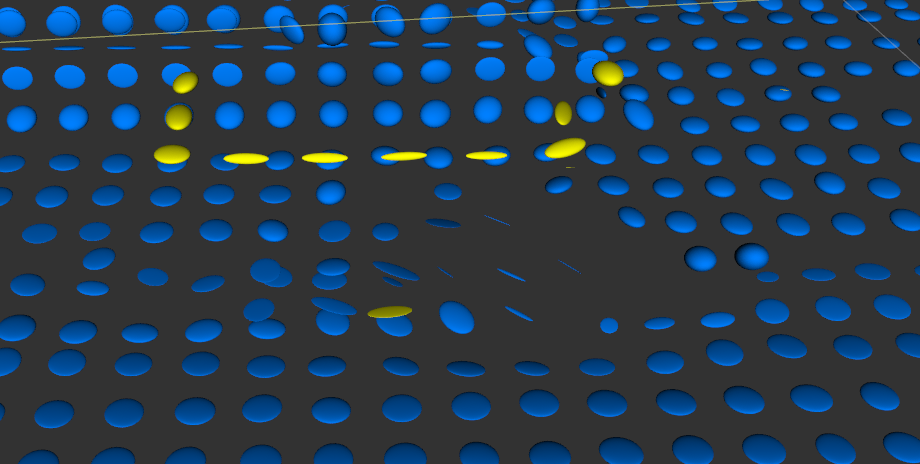}}
        \caption{NDT-OM}
        \label{fig:occlusion-om}
    \end{subfigure}
    \begin{subfigure}[t]{\divThree\textwidth}
        \centering
        \scalebox{\imageScaleThree}{\includegraphics{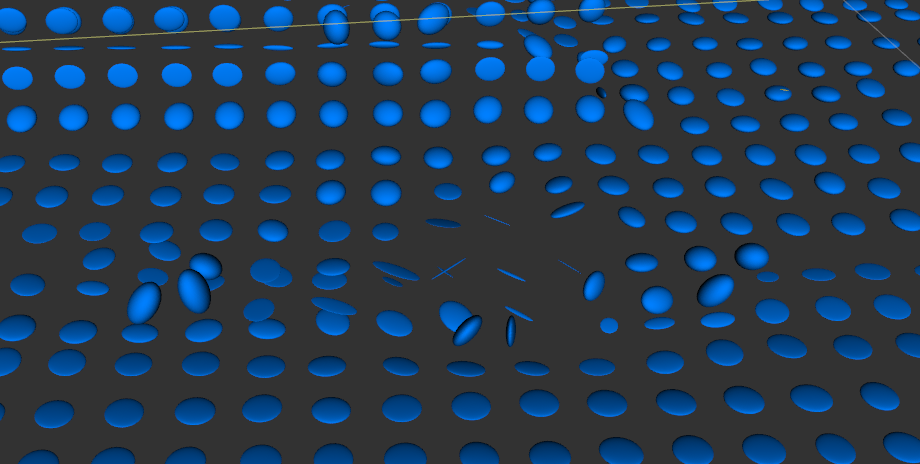}}
        \caption{Our method}
        \label{fig:occlusion-cl}
    \end{subfigure}

    \caption{The initial map (a) is updated in the lifelong scenario with NDT-OM (b) while our method (c) removes the occluded parts of the object.}
    \label{fig:qualitative-maps-occlusion}
    \figvspace{}
\end{figure}

%% file: B-qualitative-fail.tex
\newcommand\imageScaleTwo{0.175}
\newcommand\divTwo{0.155}

\begin{figure}[ht]
    \begin{subfigure}[t]{\divTwo\textwidth}
        \centering
        \scalebox{\imageScaleTwo}{\includegraphics{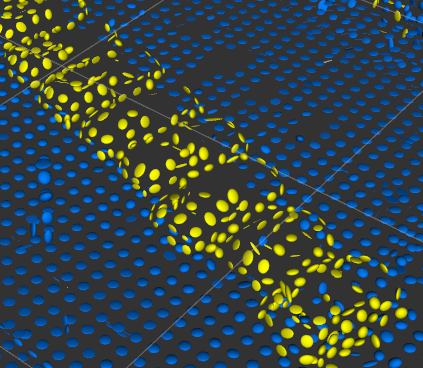}}
        \caption{Before the update}
        \label{fig:gt-lot}
    \end{subfigure}
    \begin{subfigure}[t]{\divTwo\textwidth}
        \centering
        \scalebox{\imageScaleTwo}{\includegraphics{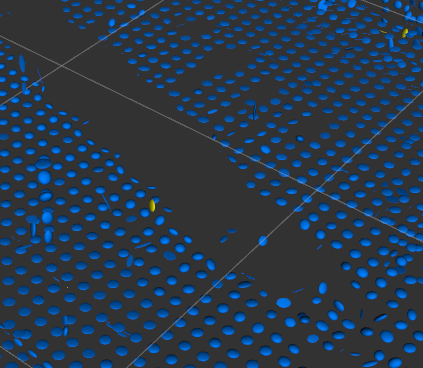}}
        \caption{our method}
        \label{fig:cl-lot}
    \end{subfigure}
    \begin{subfigure}[t]{\divTwo\textwidth}
        \centering
        \scalebox{\imageScaleTwo}{\includegraphics{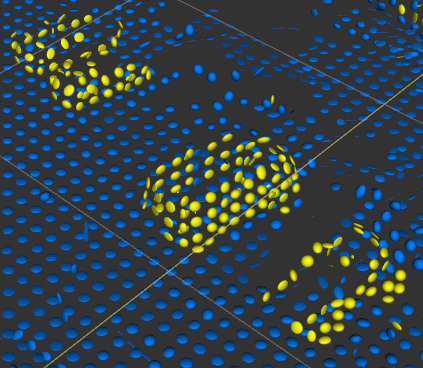}}
        \caption{Correct state after the update}
        \label{fig:d6-lot}
    \end{subfigure}

    \caption{The initial map (a) is updated in the lifelong scenario with our method (b). The correct state after the update is shown in (c). Our method erroneously clusters all of the vehicles into a single cluster and removes the whole cluster.}
    \label{fig:qualitative-maps-failure}
    \figvspace{}
\end{figure}

%% file: B-localization-results.tex
\begin{figure}
    \begin{minipage}[b]{0.55\linewidth}
        \includegraphics[width=\linewidth]{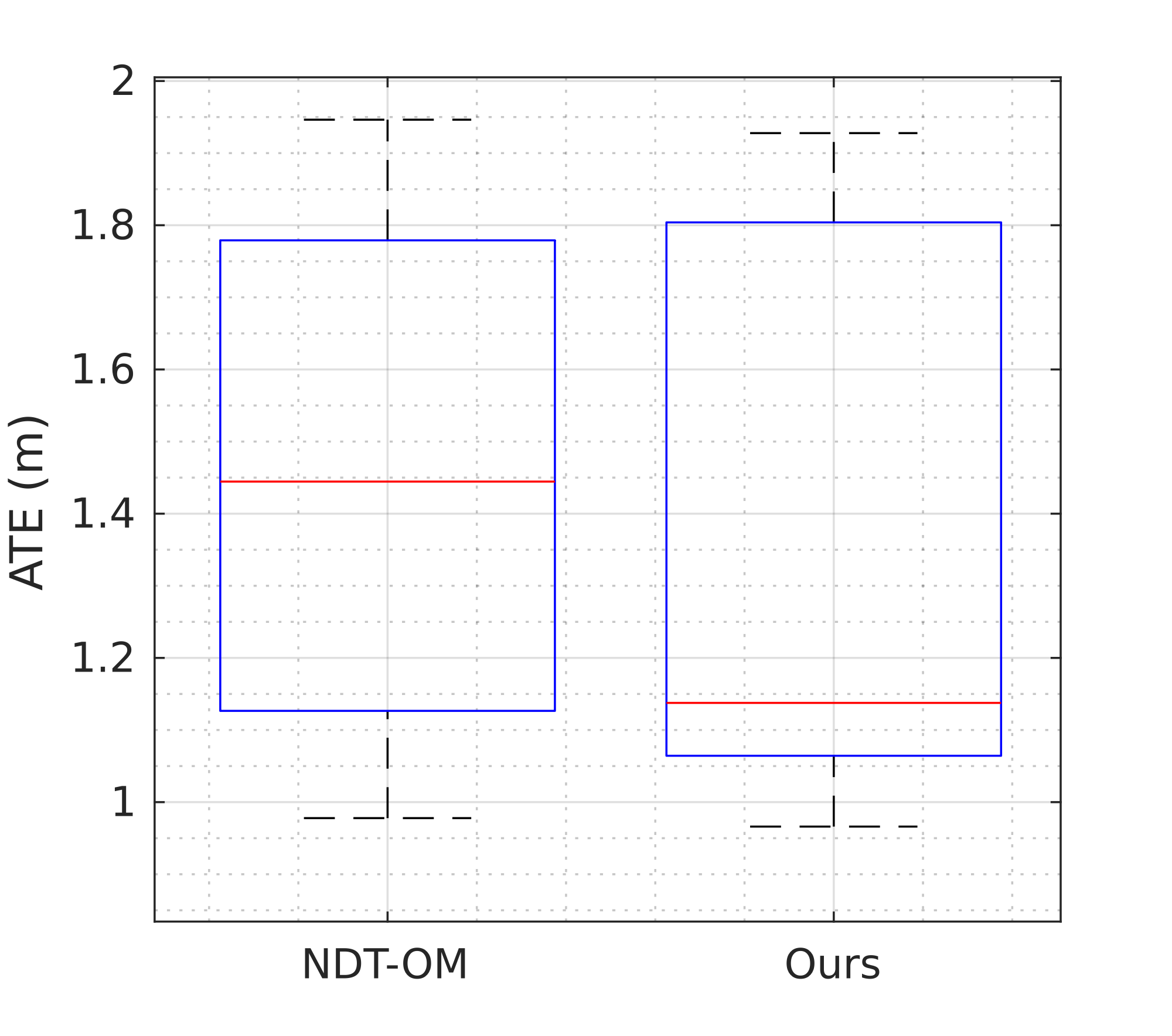}
    \end{minipage}
    \begin{minipage}[b]{0.3\linewidth}
        \centering
        \setlength\tabcolsep{4pt}
        \scalebox{0.8}{
            \begin{tabular}[b]{rrr}
                \toprule
                \textbf{Seq.} & \textbf{ \gls{ndt-om} } & \textbf{ Ours }\\
                \midrule
                2 & 1.248 m & \textbf{1.122 m} \\
                3 & 1.445 m & \textbf{0.966 m} \\
                4 & \textbf{1.573 m} & 1.579 m \\
                5 & 1.086 m & \textbf{1.045 m} \\
                6 & \textbf{0.978} m & 1.138 m \\
                7 & 1.946 m & \textbf{1.879 m} \\
                8 & \textbf{1.848 m} & 1.928 m \\
                \midrule
                mean & 1.446 m & \textbf{1.380 m} \\
                \bottomrule
            \end{tabular}
        }
        \setlength\tabcolsep{6pt}
        \vspace{0em}
    \end{minipage}
    \caption{The results of the localization experiment in terms of \gls{ate}. The red line is the sample median, the black dashed line presents the sample interval, and the blue box ranges from its first to third quartile. }
    \label{fig:localization-results}
    \figvspace{}
\end{figure}

%% file: 6-conclusion.tex
\section{Conclusion}
\label{sec:conclusion}

In this work, we formalized a relaxation of the independence of cells in grid maps by modeling the dependence of cells being occupied by the same object. This allows for more evidence to be used to estimate the cell occupancies, making the update faster and allowing the inference of the occupancy of the occluded parts of the map. Moreover, the theory allows the formalization of arbitrary dependencies between the grid cells and provides a method for updating maps using these dependencies.

The dynamic object removal capabilities of the proposed method were shown to reduce the number of residual dynamic cells in the map in both map creation and a lifelong update. This yields better-quality maps. Additionally, we studied the impact that maps built by the proposed method have on localization performance. While overall the localization gains observed are marginal, there is a slight indication that localization accuracy grows when using the proposed method. Moreover, we expect the benefits of the proposed method to go beyond localization: the residual cells left by vacated objects can be problematic in path planning; if a path is blocked, a suitable solution might be discarded needlessly. 

The study reinforces the finding that integrating geometric, semantic, and dynamic priors can improve occupancy mapping in dynamic environments. This suggests promising new research avenues in mapping, localization, and path planning with richer environment understanding.